\pdfoutput=1
\documentclass[journal,twoside,web]{ieeecolor}
\usepackage{tmi}
\usepackage{cite}
\usepackage{amsmath,amssymb,amsfonts}
\usepackage{algorithmic}
\usepackage{graphicx}
\usepackage{textcomp}
\usepackage{booktabs}
\usepackage{multirow}
\usepackage{array}
\usepackage{url}

\usepackage[hidelinks]{hyperref}

\def\BibTeX{{\rm B\kern-.05em{\sc i\kern-.025em b}\kern-.08em
    T\kern-.1667em\lower.7ex\hbox{E}\kern-.125emX}}

\begin{document}

\title{Physics-Informed Conditional Diffusion for Motion-Robust Retinal Temporal Laser Speckle Contrast Imaging}

\author{Qian~Chen,
        Yuehao~Chen,
        Qiang~Wang,
        Yutao~Feng,
        Lei~Zhu,
        and~
        Yanye~Lu
\thanks{This work was supported by the Science and Technology Innovation Key R\&D Program of Chongqing under Grant CSTB2025TIAD-STX0008 and by the Natural Science Foundation of China under Grants 82371112 and 62501020. The authors gratefully acknowledge SysEye Ltd.\ for providing hardware equipment used in this study. \textit{(Corresponding authors: Yanye Lu.)}}
\thanks{Q.~Chen, Y.~Chen, Y.~Feng, L.~Zhu and Y.~Lu are with the Department of Biomedical Engineering, Institute of Medical Technology, and National Biomedical Imaging Center, Peking University, Beijing 100871, China (e-mail: yanye.lu@pku.edu.cn).}
\thanks{Q.~Wang is with SysEye Ltd., Chongqing, China.}}

\maketitle

\begin{abstract}
Retinal laser speckle contrast imaging (LSCI) is a noninvasive optical modality for monitoring retinal blood flow dynamics. However, conventional temporal LSCI (tLSCI) reconstruction relies on sufficiently long speckle sequences to obtain stable temporal statistics, which makes it vulnerable to acquisition disturbances and limits effective temporal resolution. A physically informed reconstruction framework, termed RetinaDiff (Retinal Diffusion Model), is proposed for retinal tLSCI that is robust to motion and requires only a few frames. In RetinaDiff, registration based on phase correlation is first applied to stabilize the raw speckle sequence before contrast computation, reducing interframe misalignment so that fluctuations at each pixel primarily reflect true flow dynamics. From the long registered sequence this step yields a high-quality multiframe tLSCI map that serves only as the reconstruction target, while a motion-corrected contrast prior is computed independently from the few input frames. Next, guided by this prior, a conditional diffusion model performs inverse reconstruction by jointly conditioning on the registered few-frame sequence and the prior. On stable sequences acquired with an in-house retinal LSCI system, RetinaDiff improved SSIM from 0.159 to 0.533, PSNR from 14.83 to 18.06 dB, and FID from 211.50 to 111.55 compared with direct five-frame reconstruction, showing improved structural continuity and statistical stability over representative baselines. The framework also remains effective in a small number of extremely challenging cases, where both the direct five-frame input and the conventional multiframe reconstruction are severely degraded. Overall, this work provides a practical and physically grounded route for reliable retinal tLSCI reconstruction from extremely limited frames. The source code and model weights will be released upon acceptance.
\end{abstract}
\begin{IEEEkeywords}
Biomedical optical imaging, conditional diffusion model, few-frame reconstruction, physics-informed deep learning, retinal laser speckle contrast imaging.
\end{IEEEkeywords}

%% ============================================================
\section{Introduction}
%% ============================================================
\label{sec:introduction}

\IEEEPARstart{R}{etinal} microcirculation carries important information about ocular and systemic health, so noninvasive assessment of retinal blood flow has substantial clinical and scientific value~\cite{houben2024retinal,yuan2024retinal,balaratnasingam2023studies}. Because retinal vessels can be imaged repeatedly in vivo, retinal flow imaging also provides a useful window for studying neurovascular function and microvascular regulation. In practice, such imaging should be noncontact, repeatable, and fast enough to capture hemodynamic changes with adequate temporal resolution~\cite{han2025comparative}. Temporal laser speckle contrast imaging (tLSCI) is attractive for this purpose because it estimates perfusion from temporal fluctuations in coherent speckle patterns~\cite{zhai2024laser,li2024dual}. Compared with point scanning flowmetry or contrast agent based approaches, retinal tLSCI can support repeated perfusion monitoring without exogenous tracers~\cite{vinnett2020oct}. Its reliability, however, depends on whether the recorded speckle frames are temporally consistent and sufficiently sampled.

\begin{figure*}[!t]
    \centering
    \includegraphics[width=0.95\textwidth]{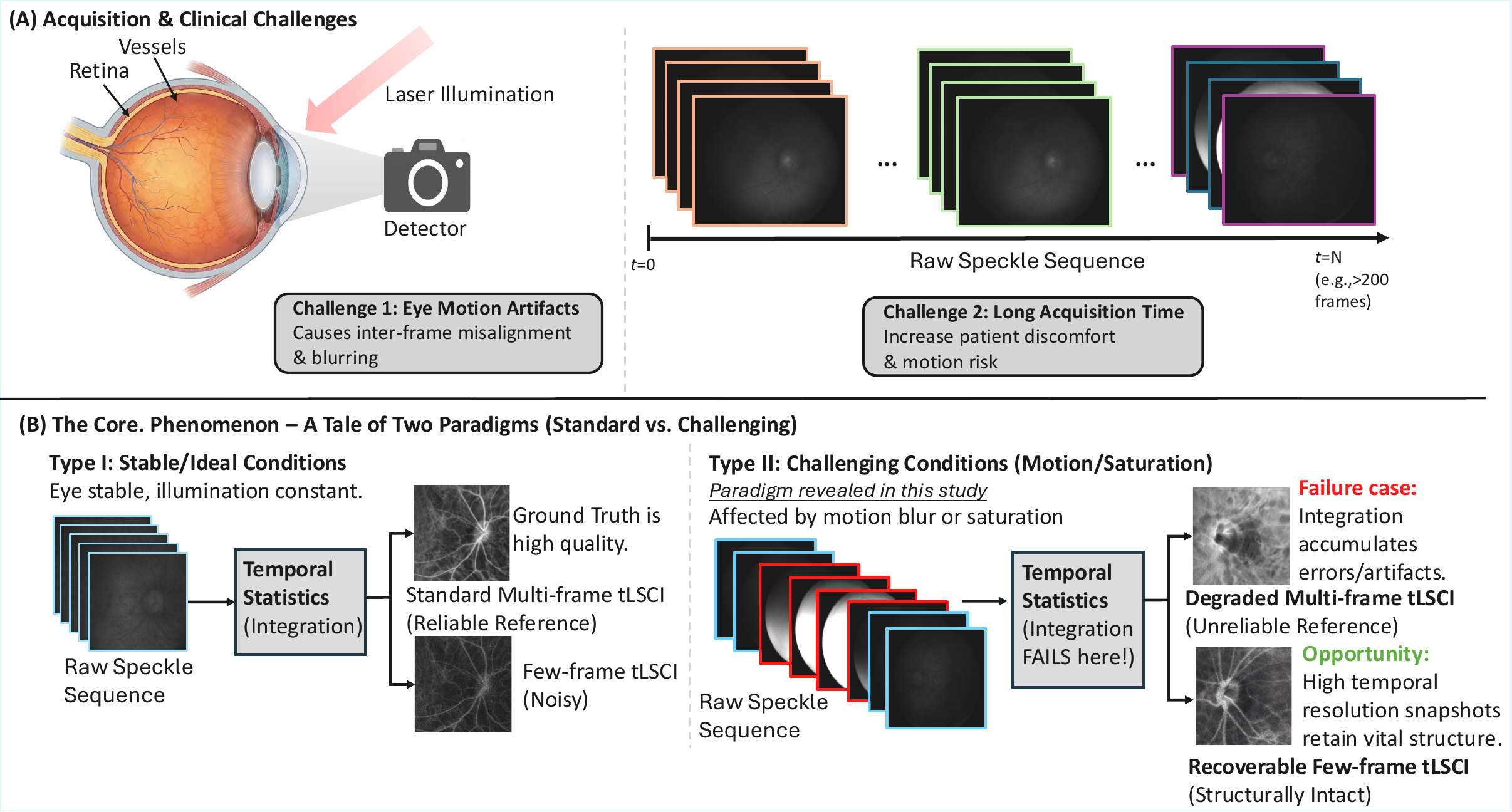}
    \caption{Problem setting of retinal tLSCI reconstruction from a small number of frames. (A) Retinal LSCI acquisition is affected by eye motion, acquisition duration, and temporal instability. (B) Under stable conditions, multiframe temporal integration provides a useful computational reference. Under challenging conditions, integration over a long temporal window may accumulate artifacts, whereas a short sequence can still retain recoverable vascular cues.}
    \label{fig:fig1}
\end{figure*}

Conventional retinal tLSCI obtains flow related contrast by estimating the temporal mean and variance of speckle intensity over a relatively long sequence~\cite{li2025research,liu2021choosing}. This strategy improves the stability of temporal statistics when the imaged tissue, illumination, and flow state remain approximately stationary. In the living eye, however, this assumption is difficult to maintain throughout an extended acquisition window. A longer window reduces statistical noise, but it also averages over a longer physiological and imaging interval. As a result, transient hemodynamic changes can be smoothed out, and the effective temporal resolution of tLSCI is reduced~\cite{feng2023sglsa}. At the same time, involuntary eye motion, illumination drift, ocular surface fluctuation, and other nonstationary disturbances can cause the intensity values at a given pixel to reflect changes in retinal position or acquisition condition rather than true local speckle decorrelation~\cite{gonzalez2023optimizing,arrigo2023quantitative,chen2024motion}. These violations of temporal stationarity can introduce artificial variance, mix vessel and background signals, and blur fine vascular structures. Consequently, long sequence retinal tLSCI may produce unstable or spatially degraded flow maps even though it uses more frames.

Reconstructing retinal tLSCI from a short sequence is therefore appealing, but it is also difficult. A small number of frames reduces the time over which motion and illumination changes can accumulate, yet it leaves too few temporal samples for reliable statistical estimation~\cite{cheng2007simplified}. This turns short sequence retinal tLSCI into a severely undersampled inverse reconstruction problem. The available frames are insufficient for direct contrast estimation, but they still contain local speckle fluctuations related to vascular flow. Prior studies have addressed parts of this problem through registration, stabilization, and refined contrast estimation~\cite{hoge2005identification,chanwimaluang2006hybrid,cheng2019manhattan,morales2021adaptive}. Phase correlation is useful in this setting because it is efficient and robust to dominant translational motion~\cite{hoge2005identification,kolar2013hybrid}. Motion compensation alone, however, cannot recover the temporal statistics lost when only a few frames are available~\cite{song2018improving}.

Learning based methods have also been used for image restoration and flow map enhancement in speckle imaging~\cite{park2024method,chen2024quantum,ahmadi2025physics}. These methods can suppress noise and improve appearance, but a purely data driven regressor may weaken the physical link between speckle fluctuations and flow related indices. For retinal tLSCI, the target is not only a visually plausible image. The reconstructed map should remain consistent with the temporal speckle formation process. Models trained only from image statistics may also be sensitive to changes in acquisition conditions, motion patterns, or image distributions~\cite{guan2021domain}.

Diffusion models provide a natural way to treat this task as conditional inverse reconstruction rather than generic image synthesis. The reconstruction should be constrained by the raw speckle observations and by physical evidence derived from temporal contrast. In particular, the inverse relationship between speckle contrast and flow related activity can narrow the solution space~\cite{briers1996laser,qiu2022inter}. A conditional diffusion model can then use this physical cue together with the observed frames to estimate a reconstruction that is both data supported and physically interpretable~\cite{wu2024fourier,kim2022diffusemorph,li2024microscopy,abbasi2025physics}.

Figure~\ref{fig:fig1} illustrates the problem setting. Under stable conditions, multiframe temporal integration provides a useful computational reference. Direct reconstruction from a few frames is usually noisy and loses fine vessel detail. Under challenging conditions, however, integration over a long temporal window may accumulate artifacts caused by motion, saturation, or illumination instability. A short sequence can still preserve recoverable vascular information in such cases. These observations motivate a reconstruction framework that can use limited temporal observations without depending unconditionally on long window statistics.

We propose RetinaDiff, a physically informed conditional diffusion framework for motion robust retinal tLSCI reconstruction from a few frames. RetinaDiff first stabilizes the raw speckle sequence with phase correlation and computes a motion corrected contrast prior. It then reconstructs the retinal tLSCI map by conditioning a diffusion model on the aligned speckle frames and the contrast prior. This design allows the reconstruction to use both local temporal evidence and global flow related structure.

Our emphasis is not to replace the physics of tLSCI with a learned image generator. Instead, we use learning to solve the part of the problem that conventional temporal statistics cannot solve when only a few frames are available. The raw frames retain the measurements, the contrast prior retains the flow related constraint, and the diffusion process provides a flexible inverse solver that can combine them under strong undersampling. This distinction is important for retinal imaging, where a visually clean map is not sufficient unless it remains tied to the acquired speckle sequence.

The main contributions are as follows.
\begin{itemize}
    \item We formulate retinal tLSCI from a few frames as a motion robust inverse reconstruction problem with severely undersampled temporal observations.
    \item We develop a dual condition strategy that combines aligned raw speckle frames with a motion corrected contrast prior for diffusion based reconstruction.
    \item We evaluate reconstruction behavior under stable, challenging, and boundary acquisition conditions, and show improved vessel continuity and statistical stability.
\end{itemize}

%% ============================================================
\section{Methods}
%% ============================================================
\label{sec:methods}

\subsection{Problem Formulation and Framework Overview}

\begin{figure*}[!t]
    \centering
    \includegraphics[width=0.95\textwidth]{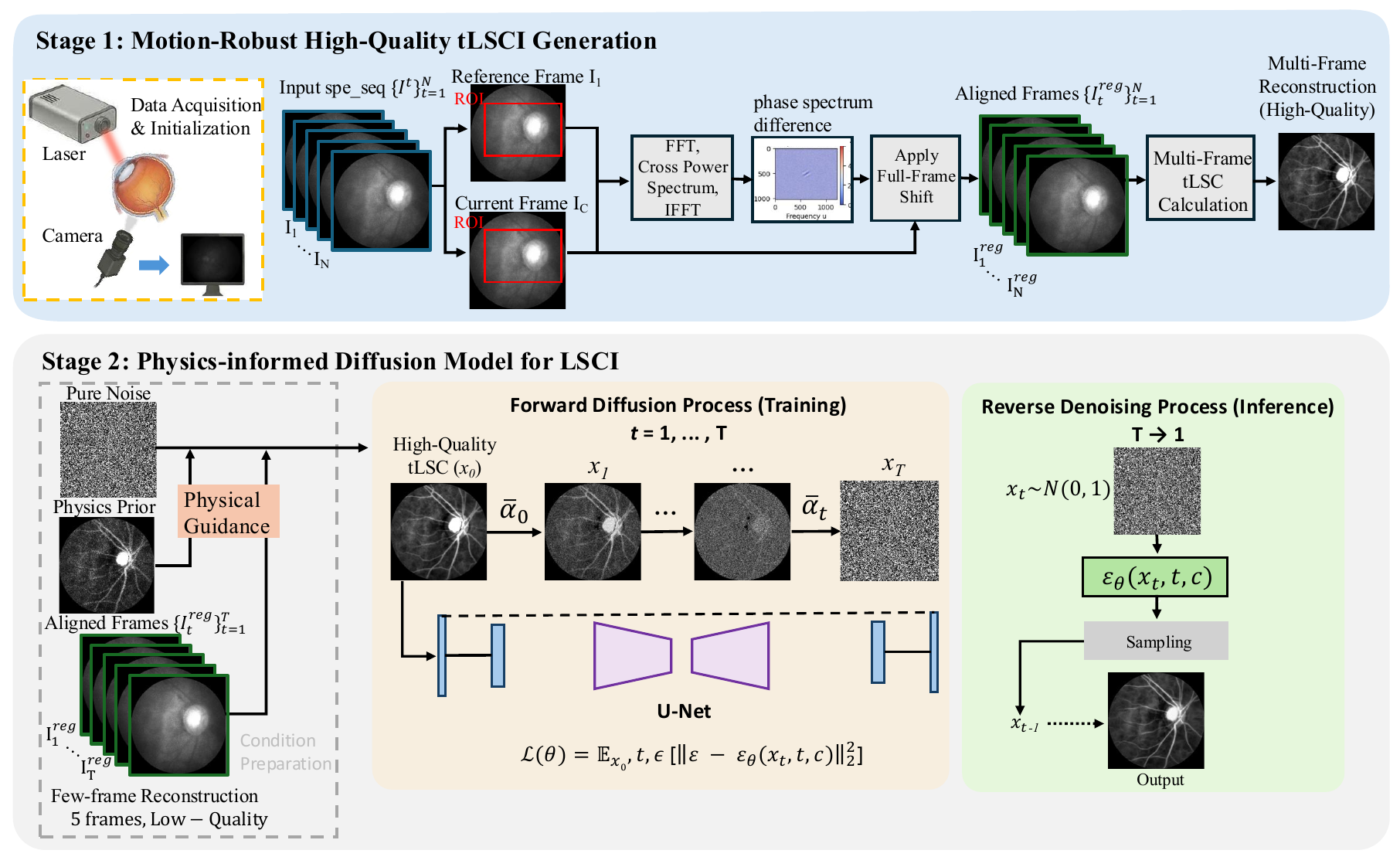}
    \caption{Overview of the proposed conditional diffusion reconstruction framework. Stage~1 stabilizes the raw speckle sequence and constructs a motion corrected contrast prior. Stage~2 reconstructs the retinal tLSCI map from a few aligned frames by conditioning the diffusion model on both raw observations and the contrast prior.}
    \label{fig:overall_framework}
\end{figure*}

Figure~\ref{fig:overall_framework} illustrates the overall framework of RetinaDiff. Given a small number of motion stabilized speckle frames, we aim to reconstruct a retinal tLSCI map that remains consistent with the long sequence reconstruction under stable acquisition. The task is formulated as an inverse reconstruction problem with limited temporal observations. In the first stage, phase correlation is used to align the raw speckle frames and reduce interframe inconsistency caused by involuntary eye motion~\cite{kolar2013hybrid,li2025phase}. Temporal speckle contrast is then computed from the long (200 frame) aligned sequence to construct the computational reference used as the training target, whereas the motion corrected contrast prior is computed separately from only the five aligned input frames~\cite{briers1996laser}. The prior therefore shares no temporal samples with the reconstruction target and introduces no information from the long sequence into the conditioning input. In the second stage, we use a conditional diffusion model to reconstruct the retinal tLSCI map from the aligned short sequence, with the contrast prior serving as physical guidance. The prior provides coarse and interpretable flow related structure, while the aligned raw frames preserve local speckle observations that are not fully captured by the prior. By combining these two sources of information, the reconstruction is constrained by both physical evidence and observed measurements, rather than being generated as an unconstrained synthetic image.

\subsection{Data Acquisition and Preprocessing}

We acquired retinal speckle sequences with an in-house retinal LSCI setup. The setup followed the general tLSCI imaging paradigm for noncontact blood flow monitoring based on temporal speckle fluctuations~\cite{feng2022functional}. A coherent near infrared laser source illuminated the retina. The source used a single mode laser diode (RLD85NZJ4-00A, ROHM, Japan) with a nominal wavelength of 850~nm. The optical power at the corneal plane was approximately 0.8~mW. A monochrome medical camera (Basler aca2440-75um med, Basler AG, Germany) recorded the backscattered speckle patterns at a raw sensor resolution of 1024 $\times$ 1024 pixels. The camera operated in free-run mode with an actual frame rate of approximately 67~fps and an exposure time close to 1/75~s. Table~\ref{tab:system_specs} summarizes the acquisition parameters.

\begin{table}[!t]
\centering
\caption{Data acquisition parameters of the retinal LSCI setup.}
\label{tab:system_specs}
\renewcommand{\arraystretch}{1.15}
\begin{tabular}{ll}
\toprule
Parameter & Value \\
\midrule
Laser diode & RLD85NZJ4-00A \\
Wavelength & 850~nm \\
Corneal power & 0.8~mW \\
Camera model & Basler aca2440-75um med \\
Raw image size & 1024 $\times$ 1024 pixels \\
Effective ROI & 768 $\times$ 768 pixels \\
Acquisition mode & Free-run \\
Exposure time & $\approx$1/75~s ($\approx$13.3~ms) \\
Frame rate & 67~fps actual / 83~fps theoretical \\
\bottomrule
\end{tabular}
\end{table}

We center cropped each raw $1024 \times 1024$ speckle image to a $768 \times 768$ region to remove edge artifacts and dark borders. The dataset contained speckle sequences from 198 independent eyes. We used 122 sequences for training and reserved 76 sequences for testing. Each training sequence was divided into four nonoverlapping $384 \times 384$ patches, yielding 488 training samples. The test set contained 8 stable sequences, 64 challenging nonstationary sequences, and 4 extremely challenging boundary cases.

For each sample, a registered 200 frame subsequence was used to construct the multiframe tLSCI reference. Five aligned frames from the same acquisition were used as the short sequence input. This paired construction reduced variability unrelated to the reconstruction task.

We used the 200 frame reconstruction as a computational reference only when the acquisition remained sufficiently stable over the temporal window. In this regime, the long sequence estimate averages enough temporal samples to provide a useful target for supervised training and quantitative evaluation. The five frame input represents a much shorter observation window and therefore tests whether the model can recover a stable tLSCI representation when direct temporal contrast estimation is strongly undersampled. For nonstationary and saturated sequences, we did not regard the long sequence result as ground truth. We used those cases to examine whether useful structure could still be recovered and to identify the point at which the measurements no longer support reliable reconstruction.

\subsection{Stage 1: Motion Stabilization and Physics Prior Construction}

\subsubsection{Motion Stabilization via Phase Correlation}

Given a raw speckle sequence $\{I_t(\mathbf{x})\}_{t=1}^{N}$, we used the first frame $I_1$ as the reference. For each subsequent frame, phase correlation estimated the displacement relative to this reference:
\begin{equation}
R_t(\mathbf{u})=
\frac{\mathcal{F}(I_t)(\mathbf{u})\,\mathcal{F}(I_1)^*(\mathbf{u})}
{\left|\mathcal{F}(I_t)(\mathbf{u})\,\mathcal{F}(I_1)^*(\mathbf{u})\right|+\epsilon},
\label{eq:phasecorr_main}
\end{equation}
\begin{equation}
\Delta_t=\arg\max_{\boldsymbol{\delta}} \mathcal{F}^{-1}(R_t)(\boldsymbol{\delta}),
\label{eq:shift_main}
\end{equation}
and the aligned frame was obtained as
\begin{equation}
\tilde{I}_t(\mathbf{x})=I_t(\mathbf{x}+\Delta_t).
\label{eq:align_main}
\end{equation}

This step compensated for dominant global translational motion before temporal contrast computation. The same registration was applied to both the long (200 frame) sequence used to reconstruct the multiframe reference and the five frame subsequence used to build the contrast prior; the two were processed independently and did not share temporal samples.

\subsubsection{Temporal Contrast Computation and Physics Prior}

We computed temporal speckle contrast from an aligned sequence as
\begin{equation}
K(\mathbf{x})=\frac{\sigma(\mathbf{x})}{\mu(\mathbf{x})+\epsilon},
\label{eq:tLSC_main}
\end{equation}
where
\begin{equation}
\mu(\mathbf{x})=\frac{1}{N}\sum_{t=1}^{N}\tilde{I}_t(\mathbf{x}),
\quad
\sigma(\mathbf{x})=
\sqrt{\frac{1}{N}\sum_{t=1}^{N}\big(\tilde{I}_t(\mathbf{x})-\mu(\mathbf{x})\big)^2}.
\label{eq:tLSC_mu_sigma}
\end{equation}
We then defined a motion corrected contrast prior from the inverse relationship between speckle contrast and flow related activity:
\begin{equation}
F_{\mathrm{phy}}(\mathbf{x})=\frac{1}{K(\mathbf{x})^2+\epsilon}.
\label{eq:phy_main}
\end{equation}
This form follows the classical tLSCI relationship in which lower speckle contrast generally corresponds to faster decorrelation and higher flow related activity under a fixed exposure setting~\cite{briers1996laser,qiu2022inter}. When constructing the prior, the contrast in~\eqref{eq:tLSC_main}--\eqref{eq:phy_main} is evaluated with $N=N_{\mathrm{few}}=5$, i.e.\ only over the five input frames; the same operator with $N=200$ is used solely to build the reference target. We did not treat this prior as an absolute flow measurement. Instead, we used it as a relative contrast derived surrogate that encodes coarse vascular structure for the diffusion model.

This distinction is important because the prior is itself estimated from a short and imperfect sequence. It can preserve the approximate vascular layout, but it may also be noisy, locally biased, or oversmoothed when the five frames contain residual motion or intensity fluctuation. We therefore used the prior as a condition rather than as a direct output. The model was required to reconcile this prior with the aligned raw speckle frames, which helped prevent the reconstruction from being determined by a single imperfect estimate.

\subsection{Stage 2: Physically Informed Conditional Diffusion Reconstruction}

Let $x_0$ denote the long sequence tLSCI reconstruction used as the training target under stable acquisition conditions. RetinaDiff reconstructs $x_0$ from a short sequence by estimating a conditional reverse process. The condition is constructed by channelwise concatenation of the aligned short sequence and the motion corrected contrast prior:
\begin{equation}
c=\mathrm{Concat}(\tilde{I}_1,\tilde{I}_2,\ldots,\tilde{I}_{N_{\mathrm{few}}},F_{\mathrm{phy}}),
\label{eq:condition}
\end{equation}
where $F_{\mathrm{phy}}$ is computed from the same short sequence using the temporal contrast model. This condition provides both coarse vascular topology from the prior and local speckle fluctuations from the raw frames. We selected this dual condition because the two inputs fail in different ways. The contrast prior summarizes temporal variation and gives the network a physically meaningful estimate of vessel layout. The raw frames retain local intensity and speckle fluctuations that may be lost during contrast computation. When the prior is locally unreliable, the raw frames can provide evidence for correction. When the raw frames are noisy or ambiguous, the prior can regularize the reconstruction toward flow related structure. In this sense, RetinaDiff uses the diffusion model as a conditional estimator rather than as an unconstrained generator.

During training, the forward diffusion process perturbs the target map by adding Gaussian noise:
\begin{equation}
q(x_t|x_0)=
\mathcal{N}
\left(
x_t;
\sqrt{\bar{\alpha}_t}x_0,
(1-\bar{\alpha}_t)\mathbf{I}
\right),
\label{eq:forward_distribution}
\end{equation}
which can be equivalently written as
\begin{equation}
x_t=\sqrt{\bar{\alpha}_t}x_0+
\sqrt{1-\bar{\alpha}_t}\epsilon,
\quad
\epsilon\sim\mathcal{N}(0,\mathbf{I}),
\label{eq:forward_sampling}
\end{equation}
where $t\in\{1,\ldots,T\}$ is the diffusion step and $\bar{\alpha}_t=\prod_{s=1}^{t}\alpha_s$ is determined by the predefined noise schedule. The denoising network $\epsilon_{\theta}$ receives the noisy target $x_t$, the diffusion step $t$, and the condition $c$. It is optimized with the standard noise prediction objective:
\begin{equation}
\mathcal{L}_{\mathrm{diff}}=
\mathbb{E}_{x_0,t,\epsilon}
\left[
\left\|
\epsilon-\epsilon_{\theta}(x_t,t,c)
\right\|_2^2
\right].
\label{eq:loss_main}
\end{equation}
This objective trains the model to denoise under explicit observational and physical conditions. For retinal tLSCI from a few frames, the condition restricts the reconstruction space by combining the contrast prior with the observed speckle fluctuations. The model is therefore encouraged to recover a long sequence consistent map while avoiding vascular structures that are unsupported by the input. The predicted clean estimate at step $t$ can be expressed as
\begin{equation}
\hat{x}_0(x_t,t,c)=
\frac{x_t-\sqrt{1-\bar{\alpha}_t}\,
\epsilon_{\theta}(x_t,t,c)}
{\sqrt{\bar{\alpha}_t}}.
\label{eq:x0_prediction}
\end{equation}

During inference, deterministic DDIM sampling reconstructs the final tLSCI map from Gaussian noise under the same condition $c$:
\begin{equation}
x_{t-1}=
\sqrt{\bar{\alpha}_{t-1}}\hat{x}_0(x_t,t,c)
+
\sqrt{1-\bar{\alpha}_{t-1}}\,
\epsilon_{\theta}(x_t,t,c),
\label{eq:ddim_sampling}
\end{equation}
which is the deterministic DDIM update, i.e.\ the special case $\eta=0$ of the general DDIM sampler in which the stochastic term vanishes. This formulation reduces the number of sampling steps while preserving structural fidelity. Before training, all target maps, contrast priors, and speckle inputs were percentile clipped and normalized to $[-1,1]$. This preprocessing reduced the influence of outliers caused by glare, static scatterers, and near saturation.

\subsection{Implementation, Baselines, and Evaluation}
All models were implemented in PyTorch and trained on a single NVIDIA GeForce RTX 3090 Ti GPU with 24~GB VRAM. RetinaDiff used a U-Net denoising backbone with 7 input channels, corresponding to $x_t$, five aligned speckle frames, and the contrast prior. The network output one channel for noise prediction. The base channel number was 64. The channel multipliers were $(1,2,4,8)$, with two residual blocks per level, attention at resolutions of 16 and 8, four attention heads, GroupNorm normalization, and SiLU activation. The diffusion model was trained with 1000 linear noise steps. We used AdamW with an initial learning rate of $1\times10^{-4}$, weight decay of $1\times10^{-2}$, a batch size of 4, and 1200 epochs. Training patches were randomly cropped to $384\times384$ pixels. Data augmentation included horizontal flipping, vertical flipping, and rotations by multiples of $90^\circ$. During inference, deterministic DDIM sampling used 100 steps with $\eta=0$.

For comparison, we evaluated Direct-5f, U-Net~\cite{ronneberger2015u}, GAN~\cite{goodfellow2014generative}, Restormer~\cite{zamir2022restormer}, NAFNet~\cite{chen2022nafnet}, and SwinIR~\cite{liang2021swinir}. All comparison methods used the same data split and preprocessing whenever applicable. For stable sequences, SSIM, PSNR, and FID were computed against the 200 frame computational reference. For challenging sequences, the multiframe output could itself be degraded, so it was used only as a comparison baseline. Direct-5f was computed from the same five aligned frames and therefore represents the conventional short window estimate without learned reconstruction. The learning based baselines were used to test whether standard restoration architectures could recover the reference from limited observations under the same data split. SSIM and PSNR measured pixel level agreement with the stable long sequence reference, while FID measured distributional similarity between reconstructed and reference maps. We report these metrics only for stable acquisitions because their interpretation depends on the reference being reliable. For challenging cases, we used visual comparison and boundary analysis, since a degraded multiframe image should not be promoted to a quantitative target.
%% ============================================================
\section{Results}
%% ============================================================
\label{sec:results}

We evaluated two acquisition regimes. The stable subset contained sequences for which the 200 frame reconstruction provided a reliable computational reference. The challenging subset contained nonstationary sequences in which motion, illumination fluctuation, or saturation degraded the conventional multiframe result. We also examined four boundary cases in which both the short sequence input and the multiframe reconstruction were severely corrupted. Unless otherwise specified, the input contained five speckle frames, and the reference was reconstructed from 200 registered frames.

\subsection{Performance under Stable Acquisition Conditions}

We first evaluated RetinaDiff under stable acquisition conditions. In these sequences, the eye remained sufficiently steady and illumination stayed approximately constant over the acquisition window. The 200 frame reconstruction therefore served as a reference for fidelity evaluation.

\begin{figure*}[!t]
    \centering
    \includegraphics[width=0.95\textwidth]{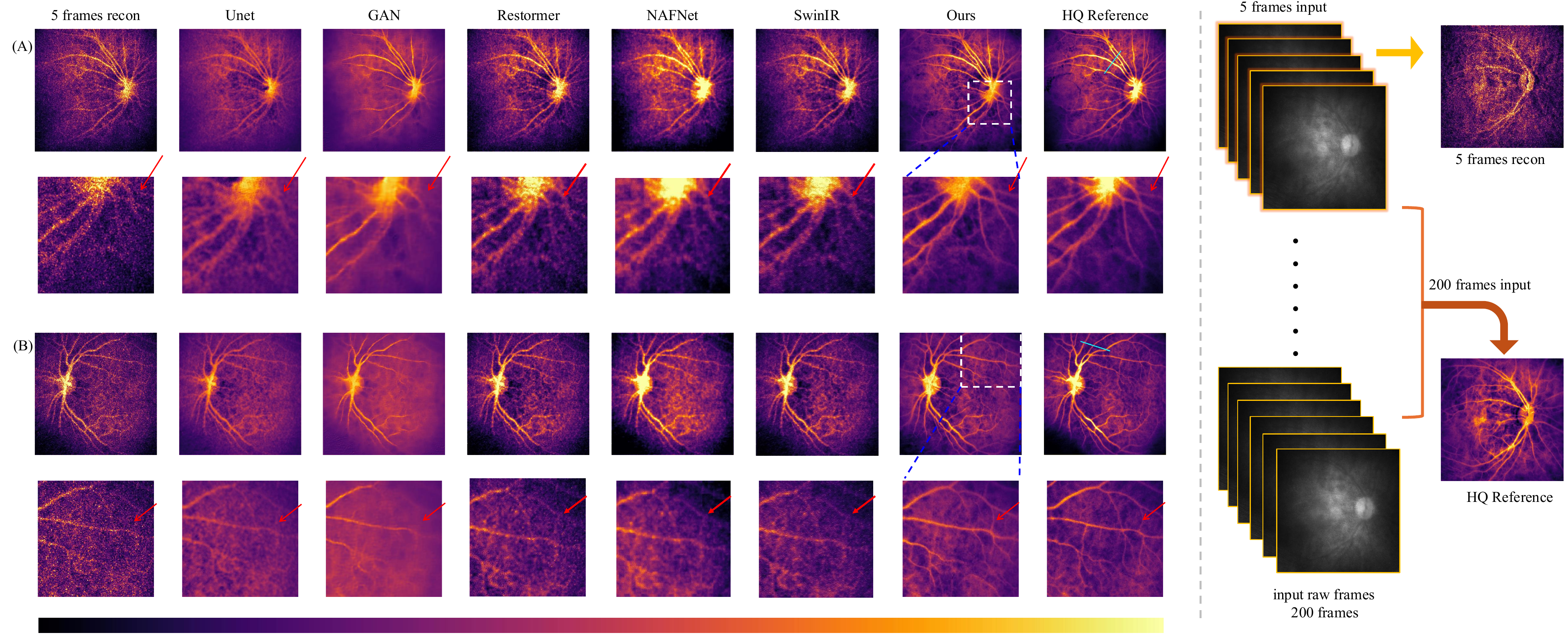}
    \caption{Representative reconstruction results under stable acquisition conditions. Columns show Direct-5f, U-Net~\cite{ronneberger2015u}, GAN~\cite{goodfellow2014generative}, Restormer~\cite{zamir2022restormer}, NAFNet~\cite{chen2022nafnet}, SwinIR~\cite{liang2021swinir}, RetinaDiff, and the long sequence computational reference. Insets highlight local vessel continuity and fine vascular details.}
    \label{fig:stable_results}
\end{figure*}

Figure~\ref{fig:stable_results} shows representative results. Direct-5f produced unstable background estimates, fragmented vessel responses, and inconsistent local contrast because the temporal statistics were severely undersampled. The supervised baselines improved global consistency, but some thin vessels were attenuated or locally oversmoothed. RetinaDiff preserved vascular topology, thin branches, and vessel continuity more closely, especially near bifurcations and peripheral low contrast regions. The enlarged regions and line profiles further show that RetinaDiff did not simply smooth the input. It recovered vessel segments that were fragmented in Direct-5f and maintained clearer vessel boundaries than U-Net and GAN. Its profiles also followed the reference more closely around vessel centers and edges.

These qualitative differences are important for retinal tLSCI because small discontinuities can change the apparent connectivity of the vascular tree. In the representative examples, the direct estimate often changed vessel appearance from one local region to another, which made low contrast branches difficult to separate from background fluctuation. RetinaDiff reduced this local inconsistency while preserving vessel transitions at branch points. The result was not a uniformly sharpened image. Instead, the reconstruction kept high contrast vessels continuous and left nonvascular background regions comparatively stable, which is consistent with the intended role of the physical condition.

Table~\ref{tab:quantitative_comparison} summarizes the quantitative results on the stable subset. RetinaDiff achieved the best SSIM, PSNR, and FID among all compared methods. Relative to Direct-5f, it improved SSIM from 0.159 to 0.533 and PSNR from 14.83~dB to 18.06~dB. These gains indicate that conditional reconstruction reduced the undersampling artifacts of short sequence tLSCI. Compared with U-Net and GAN, the numerical improvements were smaller but consistent with the visual improvements in vessel continuity and fine detail. The lower FID also indicates that the reconstructed maps better matched the distribution of the reference maps. These metrics should be interpreted as reconstruction fidelity measures rather than direct clinical endpoints.

\begin{table*}[t]
\centering
\caption{Quantitative comparison under stable acquisition conditions.}
\label{tab:quantitative_comparison}
\begin{tabular}{lccccccc}
\toprule
\textbf{Metric} & \textbf{Direct-5f} & \textbf{U-Net} & \textbf{GAN} & \textbf{Restormer} & \textbf{NAFNet} & \textbf{SwinIR} & \textbf{RetinaDiff} \\
\midrule
PSNR (dB) $\uparrow$ & 14.83 & 17.86 & 16.86 & 17.59 & 17.23 & 17.93 & \textbf{18.06} \\
SSIM $\uparrow$ & 0.159 & 0.505 & 0.529 & 0.301 & 0.381 & 0.310 & \textbf{0.533} \\
FID $\downarrow$ & 211.50 & 166.28 & 173.69 & 160.10 & 161.42 & 166.85 & \textbf{111.55} \\
\bottomrule
\end{tabular}
\end{table*}

\subsection{Performance under Challenging and Extremely Challenging Conditions}
We next evaluated a regime in which reconstruction from a long temporal window was also degraded. In our data, these cases were mainly associated with saturation from excessive instantaneous illumination, illumination nonuniformity, and motion contamination~\cite{shen2020modeling,rasta2012spectral}. Under these conditions, temporal integration can accumulate corrupted statistics. The long sequence result therefore cannot be treated as a fully reliable ground truth.

\begin{figure*}[!t]
    \centering
    \includegraphics[width=0.75\textwidth]{figures/vis2.pdf}
    \caption{Visual comparison under challenging and extremely challenging acquisition conditions. For each case, columns show Direct-5f, Degraded Multi-f, and RetinaDiff. Degraded Multi-f denotes the conventional multiframe tLSCI result from the full sequence and is shown as a degraded baseline rather than ground truth.}
    \label{fig:challenging_results}
\end{figure*}

Figure~\ref{fig:challenging_results} shows representative challenging cases. The conventional multiframe output was affected by saturation, local intensity collapse, motion contamination, or illumination instability. Direct-5f avoided some errors introduced by long window integration, but it remained noisy and difficult to interpret. In these cases, RetinaDiff produced more continuous major vessels, clearer peripheral branches, and more stable backgrounds than Direct-5f and the degraded multiframe baseline. The result shows that a short sequence can still support useful reconstruction when it retains recoverable vascular information. The contrast prior and the raw speckle frames contributed complementary evidence in this setting.

We interpreted these challenging results conservatively. The goal was not to claim that RetinaDiff can recover information that has been lost from the measurement. Instead, we asked whether the short sequence still contained enough vascular evidence to support a structurally coherent reconstruction when long window integration had become unreliable. This distinction matters because motion and saturation affect the long and short windows differently. A long window can accumulate corrupted frames, whereas a short window may avoid part of that accumulation. Conversely, if the few available frames are also strongly corrupted, the reconstruction has no reliable physical basis.

\begin{figure*}[!t]
    \centering
    \includegraphics[width=1.0\textwidth]{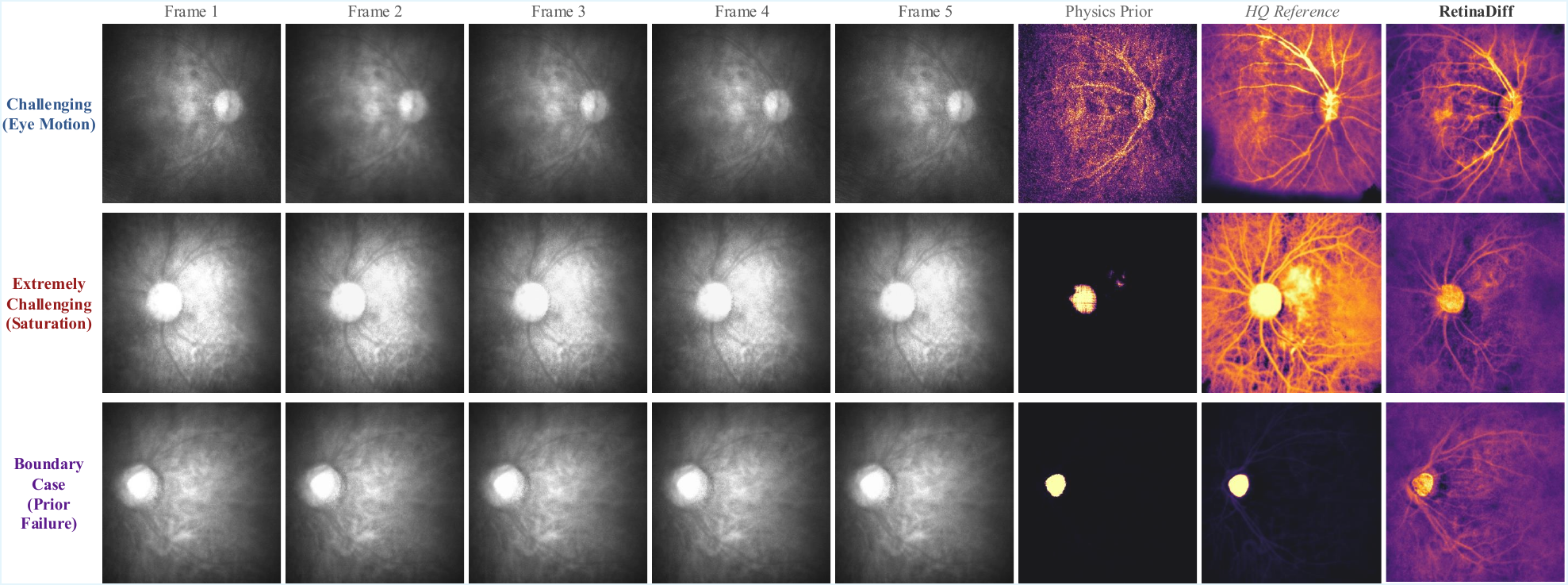}
    \caption{Row by row visualization of raw speckle frames, the contrast prior $F_{\mathrm{phy}}=1/K^{2}$, the 200 frame computational reference, and the RetinaDiff output under different acquisition conditions. As degradation increases, the prior becomes less reliable, while RetinaDiff remains effective when sufficient structural cues are still present in the raw sequence.}
    \label{fig:row_visualization}
\end{figure*}

Figure~\ref{fig:row_visualization} provides a row by row comparison of the raw frames, contrast prior, reference, and RetinaDiff output across acquisition conditions. The prior became less reliable as acquisition quality decreased. RetinaDiff remained useful when enough structural information was preserved in the short sequence. The row wise visualization also shows why the prior alone is insufficient in the degraded regime. In several examples, the prior captured the approximate vessel arrangement but amplified local artifacts caused by unstable intensity or saturation. RetinaDiff reduced some of these artifacts by using the raw frames as an additional condition. When both the prior and the raw frames were degraded in the same region, the output became less certain. This behavior is consistent with the expected information boundary of conditional reconstruction.

The extremely challenging cases should be interpreted as boundary examples rather than the primary target setting. In these cases, both the short sequence input and the multiframe reconstruction were severely degraded. This indicates that only limited recoverable information remained. RetinaDiff still produced visually plausible vascular morphology, but these cases do not prove faithful flow recovery when the input sequence is strongly corrupted.

\subsection{Ablation Analysis}

We further analyzed the contribution of each information source. We compared direct reconstruction from a few frames, reconstruction using only the contrast prior, reconstruction using only the raw input, and the full RetinaDiff model.

\begin{figure*}[!t]
    \centering
    \includegraphics[width=0.7\textwidth]{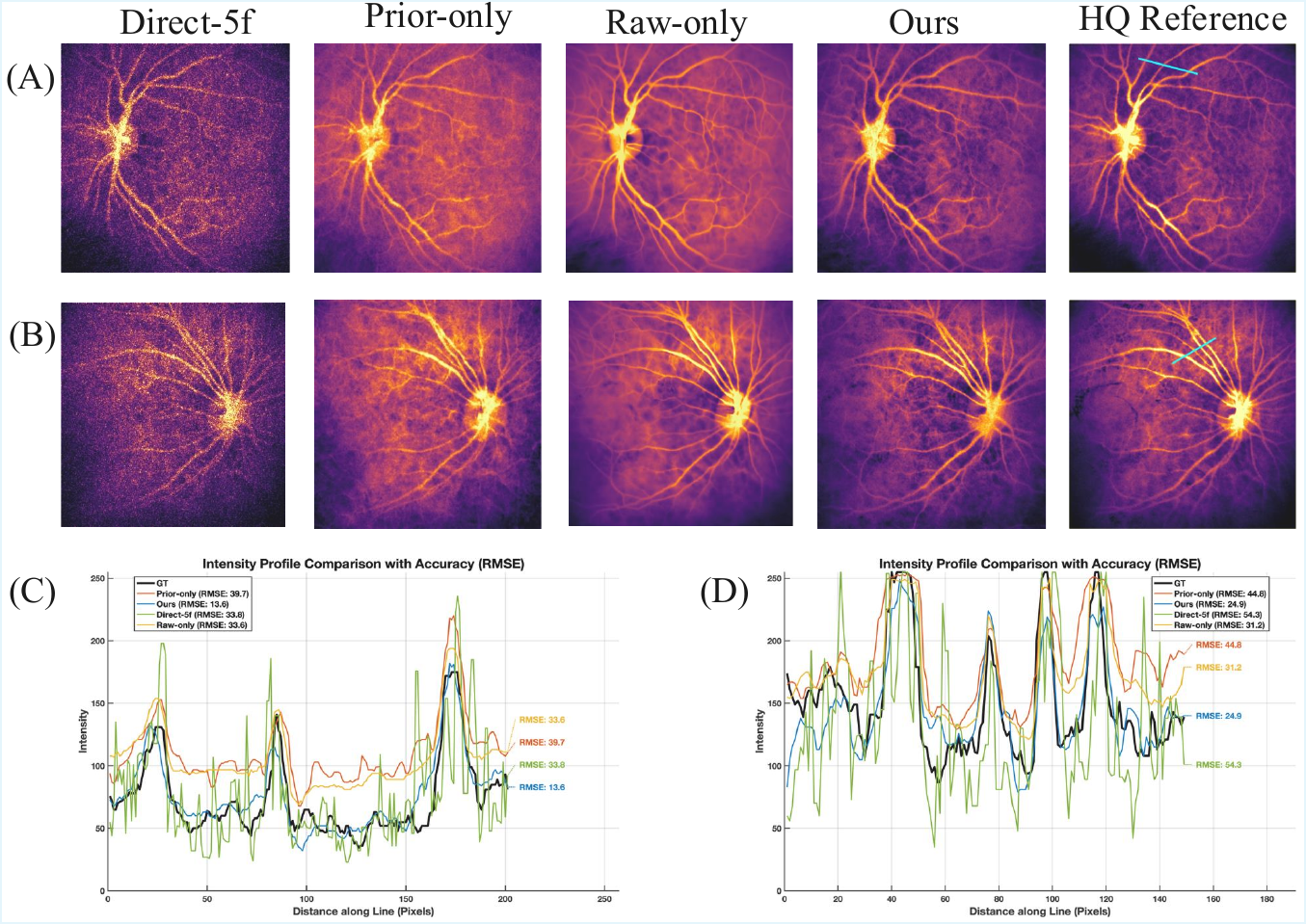}
    \caption{Ablation analysis under stable acquisition conditions. Columns show Direct-5f, Prior-only, Raw-only, RetinaDiff, and the long sequence computational reference. The contrast prior provides global vascular topology, the raw speckle frames retain local cues, and their joint use in RetinaDiff gives the closest visual agreement with the reference.}
    \label{fig:ablation_results}
\end{figure*}

Figure~\ref{fig:ablation_results} summarizes the contribution of each condition. Direct-5f was dominated by statistical fluctuation and showed fragmented vessels, which reflects the instability of temporal contrast estimation from limited samples. When we used only the motion corrected contrast prior, the reconstruction recovered the major vascular skeleton, indicating that this prior carried useful structural guidance. However, fine vessels were attenuated and local details were oversmoothed because the prior was itself derived from undersampled observations. When only the raw frames were used, more local temporal information was retained, but the reconstruction showed weaker background suppression and less consistent vascular structure without explicit physical guidance. With the full RetinaDiff model, we obtained the best balance among vascular topology, fine vessel recovery, and background stability. This result supports the dual condition design, in which the contrast prior regularizes the global flow related structure and the raw frames help refine uncertain or oversmoothed regions.

The ablation also distinguishes reconstruction fidelity from visual plausibility: the prior alone produced a smooth map that missed small branches, the raw frames alone responded to local measurements without guidance to separate flow from background, and only their combination remained physically guided while still using local speckle evidence. These results support physically informed conditioning rather than a purely image based mapping.

%% ============================================================
\section{Discussion}
%% ============================================================
\label{sec:discussion}

The experiments reveal several important properties of RetinaDiff under different acquisition conditions. We first found that motion correction is not merely a preprocessing step, but a necessary part of the reconstruction process. Temporal speckle analysis assumes that the intensity fluctuation at each pixel is measured from the same retinal location over time. Even small residual displacements can violate this assumption and introduce artificial temporal fluctuations, especially near vessel boundaries and in low signal regions. For this reason, registration is used in RetinaDiff to stabilize both the contrast prior and the reconstruction condition.

We also found that the contrast prior improves the robustness and interpretability of the reconstruction. This prior should be interpreted as a relative flow related surrogate rather than an absolute flow measurement. It reflects the inverse relationship between speckle contrast and flow dynamics under a fixed exposure setting~\cite{briers1996laser}. By using this prior as a condition, we encourage the reconstruction to remain consistent with the tLSCI formation mechanism instead of relying only on image appearance. This helps explain why RetinaDiff preserved vascular topology and suppressed ambiguous artifacts more effectively than purely data driven baselines under stable reference conditions. In this setting, the diffusion model is not used as a generic synthesis tool or a simple smoothing operator. Rather, we use it as a conditional reconstruction model guided by both physical evidence and observed speckle measurements, consistent with recent diffusion based inverse imaging and restoration methods~\cite{dou2024diffusion,croitoru2023diffusion}.

Practical retinal flow imaging often requires a compromise between statistical stability and acquisition stability. Under stable acquisition conditions, we reconstructed retinal tLSCI maps from five frames while maintaining close agreement with the multiframe reference. Under challenging conditions, long window integration often accumulated motion, saturation, or illumination artifacts over time. In several cases, however, the short sequence still retained vascular cues, and RetinaDiff produced interpretable maps when those cues were preserved. This behavior reflects a common constraint in retinal imaging: the acquisition needs enough frames for stable temporal statistics, but the eye may not remain stable throughout a long temporal window. A shorter temporal window can reduce the accumulation of nonstationary disturbances, but it also leaves fewer samples for direct contrast estimation. We designed RetinaDiff to address this tradeoff by using the short sequence as measurement evidence and the diffusion process as a physically constrained estimator. The output should therefore be viewed as a reconstruction supported by the acquired speckle frames and the contrast prior, rather than as an independent measurement of absolute physiology.

The quantitative results should be interpreted in this context. SSIM and PSNR measure agreement with the 200 frame computational reference, while FID reflects distributional similarity between reconstructed and reference maps. Although the SSIM gain over GAN was modest, the improvement in PSNR and the reduction in FID were consistent with the visual improvements in vessel continuity, fine vessel recovery, and artifact suppression.

The boundary cases clarify the fundamental limit of this reconstruction problem. When saturation or severe instability persists across the acquisition, the temporal statistics needed to build the contrast prior can be irreparably corrupted. In such cases, the model may still generate vascular patterns, but their reliability is limited by the information preserved in the raw speckle sequence. We therefore regard these cases as examples of the recoverable boundary rather than as evidence of faithful flow recovery. The recoverable regime is bounded not only by model capacity, but also by physical observability in the acquired data.

This limitation also explains why the challenging subset should be evaluated differently from the stable subset. When the multiframe reference is itself degraded, high numerical agreement with that reference does not necessarily indicate a better reconstruction. For these stress cases, visual and structural assessment is more informative, although it remains less definitive than a calibrated flow measurement. Future validation should include independent physiological or multimodal references so that reconstruction quality can be assessed beyond agreement with a computational tLSCI target. Several limitations should be noted. In this study, we mainly corrected dominant translational motion, while more complex retinal deformation or torsional eye motion may require more advanced registration strategies. We also reconstructed relative retinal tLSCI maps rather than absolute quantitative flow. The outputs should therefore be interpreted as flow related contrast representations, and absolute perfusion quantification would require physiological calibration or validation with an independent reference. The current data were acquired with a single retinal LSCI setup, so generalization across devices, populations, and disease conditions remains to be evaluated. Although deterministic DDIM sampling reduced the number of reverse steps, diffusion inference is still more computationally demanding than direct feed forward regression.

%% ============================================================
\section{Conclusion}
%% ============================================================
\label{sec:conclusion}
In the stable subset, we found that RetinaDiff produced retinal tLSCI maps that were closer to the reference reconstructed from 200 registered frames. The reconstructed maps also preserved vessel continuity, background stability, and fine vascular detail better than direct reconstruction from a few frames and the comparison models. In the challenging subset, long window integration often introduced artifacts when motion, saturation, or illumination instability accumulated over time. When the raw frames still retained usable vascular information, however, RetinaDiff recovered interpretable tLSCI maps from the short sequence. Together, these results show that physically informed conditional diffusion can support retinal blood flow imaging with shorter acquisition windows. They also define an important practical boundary: reliable reconstruction is no longer guaranteed once the relevant speckle information has been removed by severe saturation, motion, or illumination instability. Future work will extend this study to larger clinical cohorts, improve motion modeling, evaluate performance across imaging devices, explore adaptive acquisition strategies, and accelerate inference toward real time retinal hemodynamic imaging.
%% ============================================================
%% Back matter
%% ============================================================

\bibliographystyle{IEEEtran}
\bibliography{references}

\end{document}